%% file: IEEE-conference-template-062824.tex
\documentclass[conference]{IEEEtran}
\IEEEoverridecommandlockouts

\usepackage{cite}
\usepackage{amsmath,amssymb,amsfonts}
\usepackage{algorithmic}
\usepackage{graphicx}
\usepackage{textcomp}
\usepackage{xcolor}
\def\BibTeX{{\rm B\kern-.05em{\sc i\kern-.025em b}\kern-.08em
    T\kern-.1667em\lower.7ex\hbox{E}\kern-.125emX}}

\usepackage{tabularray}
\usepackage{dsfont} 
\input{tables.tex}
\usepackage{stfloats}
\usepackage{cuted}
\begin{document}

\title{Least-Ambiguous Multi-Label Classifier
}

\author{
\IEEEauthorblockN{
  1\textsuperscript{st} Misgina Tsighe Hagos\IEEEauthorrefmark{1}\IEEEauthorrefmark{2},
  2\textsuperscript{nd} Claes Lundström\IEEEauthorrefmark{1}\IEEEauthorrefmark{2}\IEEEauthorrefmark{3}
}
\IEEEauthorblockA{\IEEEauthorrefmark{1}\textit{Department of Science and Technology}, \textit{Linköping University}}
\IEEEauthorblockA{\IEEEauthorrefmark{2}\textit{Center for Medical Imaging Science and Visualization}, \textit{Linköping University}}
\IEEEauthorblockA{\IEEEauthorrefmark{3}\textit{Sectra AB}}
}

\maketitle

\begin{abstract}

Multi-label learning often requires identifying all relevant labels for training instances, but collecting full label annotations is costly and labor-intensive. In many datasets, only a single positive label is annotated per training instance, despite the presence of multiple relevant labels. This setting, known as single-positive multi-label learning (SPMLL), presents a significant challenge due to its extreme form of partial supervision. We propose a model-agnostic approach to SPMLL that draws on conformal prediction to produce calibrated set-valued outputs, enabling reliable multi-label predictions at test time. Our method bridges the supervision gap between single-label training and multi-label evaluation without relying on label distribution assumptions. We evaluate our approach on 12 benchmark datasets, demonstrating consistent improvements over existing baselines and practical applicability.

\end{abstract}


\section{Introduction}
\label{section:introduction}

Most classification tasks in machine learning and deep learning focus on the single-label setting, where each input instance is assumed to belong to exactly one category from a predefined set~\cite{schapire2000boostexter}. However, many datasets are inherently multi-label~\cite{zhang2013review,wu2019tencent}. For example, an outdoor image may simultaneously correspond to labels such as \texttt{tree}, \texttt{rock}, and \texttt{bike-path}, rather than a single label. In such cases, training a model to accurately identify all relevant ground-truth labels is essential. This typically requires collecting a training dataset that contains all the positive labels for each instance, from human annotators, which can be costly and labor-intensive~\cite{deng2014scalable}.

We address the multi-label learning problem under the constraint that only a single positive label is observed per training instance. This setting arises naturally in domains such as medical imaging, where expert annotations are expensive and often limited. Although multiple labels may be relevant for a given instance, annotation constraints result in only one being recorded, creating an extreme case of partial supervision. Nevertheless, producing accurate multi-label predictions at test time remains critical.

We study this single-positive multi-label learning (SPMLL) problem~\cite{cole2021multi}, where each training instance is annotated with only one positive label, while evaluation requires recovering the complete set of relevant labels. We draw on ideas from calibrated, set-valued prediction in conformal prediction to reject individual label outputs in multi-label classifiers, enabling us to construct reliable model-agnostic multi-label predictors. We validate our method on 12 benchmark datasets, demonstrating strong performance and practical usability.

\section{Problem Statement}
\label{section:problem_statement}

Let $\mathcal{X} \subseteq \mathbb{R}^d$ denote the input space and $\mathcal{Y} = \{0,1\}^K$ the multi-label output space over $K$ labels. In the standard multi-label learning (MLL) setting~\cite{gibaja2015tutorial}, the training dataset consists of $N$ fully labeled examples $\{(x^{(n)}, Y^{(n)})\}_{n=1}^N$, where $x^{(n)} \in \mathcal{X}$ and $Y^{(n)} = (Y_1^{(n)}, \ldots, Y_K^{(n)}) \in \mathcal{Y}$ is a binary label vector. Here, $Y_i^{(n)} = 1$ indicates the presence of label $i$ for instance $x^{(n)}$.

The goal in MLL is to learn a predictive function $f: \mathcal{X} \rightarrow [0,1]^K$, such that for any previously unseen test input $x_{\text{test}}$, it returns a vector $f(x_{\text{test}})$ of relevance scores, where $f_i(x_{\text{test}}) \in [0,1]$ assigns a relevance score to label $i$.

The training formulation in MLL assumes that each training instance $x^{(n)}$ is associated with a fully observed label vector, where the presence and absence of every label is explicitly specified~\cite{gibaja2015tutorial}. In contrast, in SPMLL~\cite{cole2021multi}, each training instance $x^{(n)}$ is annotated with a \emph{single} positive label $y^{(n)} \in \{1, \ldots, K\}$. All other potentially relevant labels are unobserved. Thus, the training data comprises of pairs $\{(x^{(n)}, y^{(n)})\}_{n=1}^N$, where $y^{(n)}$ provides only a single positive label for $x^{(n)}$. This makes standard multi-label learning approaches ineffective, motivating the need for methods that can generalize from single-positive training to multi-label inference.

Current methods lack a model-agnostic mechanism to abstain from predicting low-confidence individual labels, which is essential in high-stakes applications. Our work addresses this gap by developing a post-hoc, model-agnostic approach to rejecting less confident individual labels in SPMLL.

\section{Related work}
\label{section:related_work}

In this section, we highlight key methods relevant to our approach on learning from single-positive labels and reject options in multi-label learning.

\subsection{Single-Positive Multi-Label Learning}

Binary cross-entropy (BCE) loss is the most commonly used loss in multi-label learning~\cite{nam2014large,veit2017learning}. For instances with fully observed labels ($x^{(n)}, Y^{(n)}$) and a model output $f(x^{(n)})$ given by $f_n$, BCE is given by,

\begin{eqnarray}
\mathcal{L}_{\mathrm{BCE}}(f_n, Y^{(n)}) = -\frac{1}{K} \sum_{i=1}^{K} \big[ \mathds{1}_{[Y_i^{(n)}=1]} \log(f_{ni}) \nonumber \\
\quad 
+ \mathds{1}_{[Y_i^{(n)}=0]} \log(1 - f_{ni}) \big]
\end{eqnarray}

\noindent where $\mathds{1}_{[Y_i^{(n)}=1]}$ and $\mathds{1}_{[Y_i^{(n)}=0]}$ represent $P(Y_i = 1 \mid x^{(n)})$ and $P(Y_i = 0 \mid x^{(n)})$, respectively. For the SPMLL case, where we have training instances ($x^{(n)}, y^{(n)}$) with a single positive label per instance, the obvious modification to the BCE loss is to treat the unobserved labels as negative. This is referred to as assume negative (AN)~\cite{cole2021multi}, and it is formulated as follows,

\begin{eqnarray}
\label{equation:loss_an}
\mathcal{L}_{\mathrm{AN}}(f_n, y^{(n)}) &=& -\frac{1}{K} \sum_{i=1}^{K} \big[ \mathds{1}_{[y_i^{(n)} = 1]} \log(f_{ni}) \nonumber \\
&& +\ \mathds{1}_{[y_i^{(n)} \neq 1]} \log(1 - f_{ni}) \big]
\end{eqnarray}

AN introduces false negatives since it treats the rest of the unobserved labels as negative. One way to circumnavigate this is to downweight the second part of the loss function in \eqref{equation:loss_an} using a weighing parameter $\gamma = \frac{1}{K-1}$, referred to as weak assume negative (WAN)~\cite{cole2021multi},

\begin{eqnarray}
\mathcal{L}_{\mathrm{WAN}}(f_n, y^{(n)}) &=& -\frac{1}{K} \sum_{i=1}^{K} \big[ \mathds{1}_{[y_i^{(n)} = 1]} \log(f_{ni}) \nonumber \\
&& +\ \mathds{1}_{[y_i^{(n)} \neq 1]} \gamma \log(1 - f_{ni}) \big]
\end{eqnarray}

\noindent where $\gamma$ ensures that the single positive has the same influence as the rest of the $K-1$ labels that are assumed negative.

AN and WAN approaches to SPMLL treat the unobserved labels as negative. Other SPMLL methods intend to recover the unobserved labels. These include regularized online label estimation (ROLE)~\cite{cole2021multi}, which estimates the unobserved labels during model training, and single positive multi-label learning with label enhancement (SMILE)~\cite{xu2022one}, which recovers the posterior density of the latent soft labels as label enhancement.  Another interesting direction in the literature reduces annotation cost by collecting complementary labels, which are known to be irrelevant to an instance \cite{gao2025multi}. MLL approaches designed to handle missing labels can also be applied to SPMLL. These include multi-label learning with global and local label correlation (GLOCAL)~\cite{zhu2017multi}, multi-label learning with missing labels (MLML)~\cite{wu2014multi}, and discriminant multi-label learning (DM2L)~\cite{ma2021expand}. GLOCAL, which is designed to handle both fully labeled and missing label scenarios, simultaneously exploits global and local label correlations by learning a latent label representation. MLML recovers complete label assignments for each sample by enforcing consistency with available label assignments and promoting label smoothness. DM2L learns from missing labels by imposing low-rank constraints on predictions of instances sharing the same labels (local structure) and encouraging a maximally separated structure for predictions of instances with different labels (global structure). The above methods usually treat inference time model outputs, $f_i(x_{\text{test}})$, equally and assume each output is sufficiently reliable. This assumption rarely holds in practice, as confidence can vary significantly across labels, especially in high-dimensional or imbalanced settings.

\subsection{Reject Options in Multi-Label Learning}

Introducing a label-wise reject option allows the model to abstain from committing to uncertain positive or negative assignments, improving robustness in high-risk applications. Existing reject option methods in multi-label learning, however, focus on instance-wise abstention, where the model refrains from predicting any labels for uncertain test inputs~\cite{pillai2011classification,pillai2013multi,wang2021can}. These do not provide fine-grained confidence estimates per label. 

Label prioritization can be incorporated into loss computation during model training. In \cite{duarte2021plm} and \cite{chen2022extreme}, for example, labels are stochastically masked during training, balancing the influence of each class. \cite{peng2023novel} employ a dimensionality reduction approach to reduce the label space for a lower computational complexity. These approaches require modifying the model training process and access to a fully annotated multi-label training dataset that is unavailable in the SPMLL setting.

\section{Proposed Method}

Our approach, which we refer to as Least Ambiguous Multi-Label Classifier (LAMC), is inspired by set-valued classification methods and adapts similar thresholding principles to the multi-label setting~\cite{sadinle2019least,angelopoulos2023conformal}. Let $f_i(x) \in [0,1]$ denote the model's predicted confidence (the sigmoid output) for class $i \in \{1, \dots, K\}$ on input $x$. To enable selective prediction with a controlled reliability, LAMC computes class-specific confidence thresholds using a held-out calibration set $\mathcal{D}_{\text{cal}}$, which is previously unseen by the model, and represents the expected test distribution. 

To compute the class-specific thresholds for each class $i$, we first collect the predicted confidence scores over $n_{cal}$ calibration examples where the class is present:
\begin{eqnarray}
S_i = \left\{ f_i(x_j) \mid (x_j, y_j) \in \mathcal{D}_{\text{cal}},\ y_{ji} = 1 \right\}.
\end{eqnarray}

We then compute a class-specific threshold $\hat{q}_i$ as the $\left\lceil (1 - \alpha)(n_{\text{cal}} + 1)/n_{\text{cal}} \right\rceil$-empirical quantile of $S_i$, where $\alpha \in (0,1)$ is a user selected error rate and controls how selective the filtering is and $(n_{\text{cal}} + 1)/n_{\text{cal}}$ is a correction for small calibration sets.

\begin{eqnarray}
    \hat{q}_i = \operatorname{quantile} \left( S_i; \left\lceil \frac{(1 - \alpha)(n_{\text{cal}} + 1)}{n_{\text{cal}}} \right\rceil \right)
\end{eqnarray}

\noindent The threshold $\hat{q}_i$ is set so that at least a $1 - \alpha$ fraction of test predictions with scores above $\hat{q}_i$ are expected to be correct for class $i$.

Let $\{x^{(j)}_{\text{test}}\}_{j=1}^N$ denote the test set. During inference, for each class $i$, we identify the indices of test inputs whose predicted scores exceed the threshold:

\begin{eqnarray}
\mathcal{I}_i = \left\{ j \mid f_i(x^{(j)}_{\text{test}}) > \hat{q}_i \right\}.
\end{eqnarray}

We then retain only these selected predictions for class $i$ as the model outputs:
\begin{eqnarray}
\left\{ f_i(x^{(j)}_{\text{test}}) \mid j \in \mathcal{I}_i \right\}.
\end{eqnarray}

Predictions for class $i$ that do not satisfy the threshold (i.e., $j \notin \mathcal{I}_i$) are \textit{excluded} from the final output set. This procedure allows the model to abstain from uncertain individual label predictions, improving the trustworthiness of the accepted predictions by leveraging class-conditional calibration.

\begin{table*}[t]
    \centering
    \TableDatasetStat
\end{table*}

\section{Experiments}

Here, we present details of our experiments' configurations, results, and discussion.

\subsection{Experiment Settings}

For easier comparison, we follow~\cite{xu2022one} and use a two-layer multi-layer perceptron (MLP). We similarly train them using an Adam optimizer for 25 epochs, with learning rates selected from the set $\{10^{-4}, 10^{-3}, 10^{-2}\}$. We use a WAN loss and a batch size of 16. All models were trained and evaluated in five runs with different initializations. We set $\alpha=0.5$ for computing the class-specific thresholds in LAMC. We use the multi-label metrics average precision (the larger the better), coverage error (the smaller the better), and ranking loss (the smaller the better) for evaluation. We compare our approach to seven methods introduced in Section~\ref{section:related_work}: SMILE, AN, WAN, ROLE, GLOCAL, MLML, and DM2L.

We use 12 datasets from~\cite{xu2022one} \footnote{https://github.com/palm-ml/smile}. Details of the datasets are presented in Table~\ref{table:dataset_stat}. For the baseline methods, we follow an 80\%/10\%/10\% train/validation/test split, whereas for our approach, we use a 70\%/10\%/10\%/10\% train/calibration/validation/test split. In both cases, the train set contains single-positive labels, while all the other sets contain fully observed labels. We use ten instances per label for the calibration set, and we use the rest to assess whether increasing the calibration set brings a performance gain.

\subsection{Experiment Results and Discussion}

\begin{table*}
    \centering
    \TableAveragePrecision
\end{table*}

\begin{table*}
    \centering
    \TableCoverageError
\end{table*}

\begin{table*}[h]
    \centering
    \TableRankingLoss
\end{table*}

\begin{figure*}
\centerline{\includegraphics[width=\textwidth]{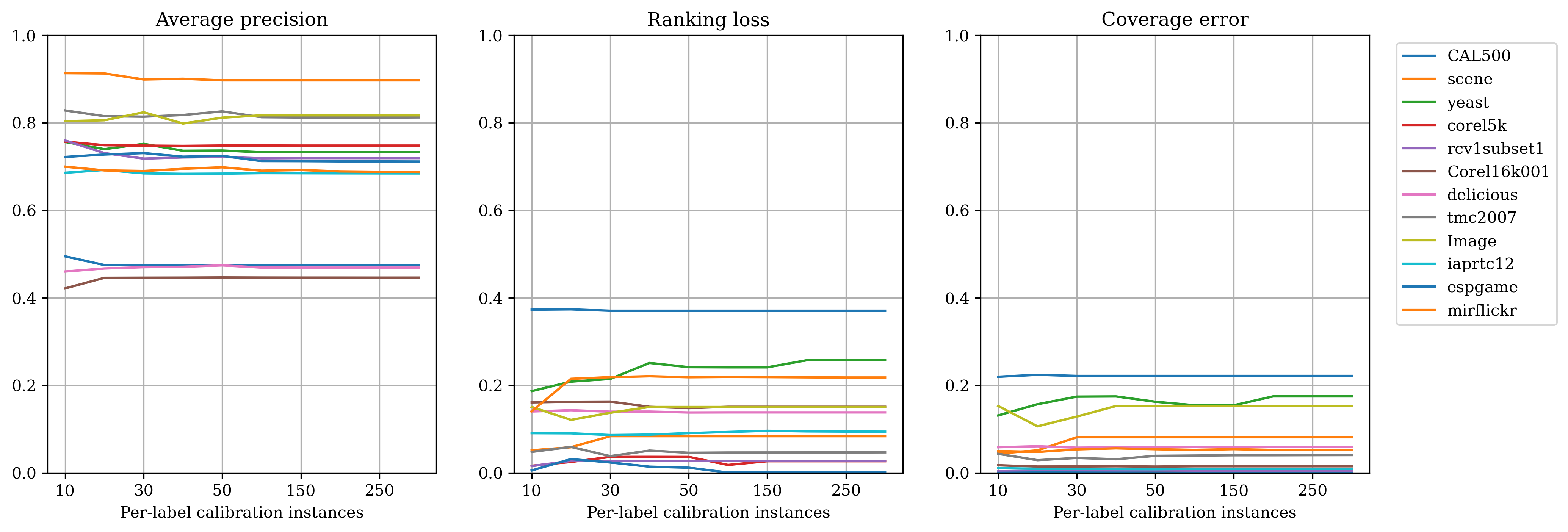}}
\caption{Effect of per-label calibration set size on the average precision, ranking loss, and coverage error metrics.} 
\label{figure:lamc_at_different_calibration_sizes}
\end{figure*}

We present the $\texttt{MEAN} \pm \texttt{STD}$ over five training runs of the average precision, coverage error, and ranking loss in Tables~\ref{table:average_precision},~\ref{table:coverage_error}, and~\ref{table:ranking_loss}. Fig.~\ref{figure:lamc_at_different_calibration_sizes} presents the performance of the LAMC approach at different per-label calibration set sizes.

LAMC outperforms existing approaches on all evaluation metrics and all datasets, as seen in Tables~\ref{table:average_precision} and~\ref{table:coverage_error}, except for the ranking loss. At ranking loss, LAMC still achieves superior performance on five datasets and achieves comparable performance to the rest of the methods on the rest of the datasets. However, as most methods already achieve strong performance on the ranking loss, there is little room for improvement.

High label cardinality negatively impacts performance. This effect is particularly evident on the CAL500 dataset, which exhibits the highest cardinality. While the average precision of most of the methods on the CAL500 is comparable to that on the other datasets, the impact of cardinality becomes more pronounced when considering ranking loss and coverage error. Nearly all the methods perform significantly worse on these metrics, highlighting the challenge in accurately ranking relevant labels under high cardinality conditions.

We find that using 10 instances per label as a calibration set is enough to generate superior performance with LAMC. The effect of an increasing calibration set is minimal. This is reported in Fig.~\ref{figure:lamc_at_different_calibration_sizes}. For all the datasets, we see that the change in evaluation metrics is limited and inconsistent as the calibration set size increases. Surprisingly, sometimes a bigger calibration set can introduce noise and reduce performance. This can be observed with the sometimes decreasing average precision, and increased ranking loss and coverage error in Fig.~\ref{figure:lamc_at_different_calibration_sizes}. Recall that larger average precision, and smaller ranking loss and coverage error are preferred.

\section{Conclusion}
\label{section:conclusion}

This paper introduces a practical approach to the single-positive multi-label learning (SPMLL) problem by leveraging class-wise calibration for a set-valued prediction. We propose the Least-Ambiguous Multi-label Classifier (LAMC), which selectively rejects low-confidence labels to produce reliable multi-label outputs under extreme partial supervision. Empirical results on 12 benchmark datasets demonstrate that LAMC consistently outperforms existing baselines, highlighting its practical effectiveness and robustness. While our method is model-agnostic and requires no distributional assumptions, it is supervised using a calibration set. Future work will consider unsupervised approaches that are independent of the dataset during calibration.

\section*{Acknowledgments}

This work was partially supported by the Wallenberg AI, Autonomous Systems and Software Program (WASP) funded by the Knut and Alice Wallenberg Foundation.

\bibliographystyle{IEEEtran}
\bibliography{refs}

\end{document}

%% file: tables.tex
\newcommand{\TableCoverageError}{
    \caption{Coverage error ($\downarrow$). Mean values and standard deviation over five runs are reported.} 
    \begin{tblr}{
      hline{1,14} = {-}{0.08em},
      hline{2} = {-}{0.05em},
    }
    Datasets    & LAMC                   & SMILE       & AN          & WAN         & ROLE        & GLOCAL      & MLML        & DM2L        \\
    CAL500      & \textbf{0.201 $\pm$ 0.017} & 0.850$\pm$0.018 & 0.891$\pm$0.009 & 0.828$\pm$0.046 & 0.942$\pm$0.024 & 0.882$\pm$0.012 & 0.708$\pm$0.002 & 0.720$\pm$0.012 \\
    scene       & \textbf{0.055$\pm$0.015} & 0.086$\pm$0.020 & 0.205$\pm$0.092 & 0.185$\pm$0.021 & 0.177$\pm$0.007 & 0.154$\pm$0.005 & 0.404$\pm$0.011 & 0.905$\pm$0.009 \\
    yeast       & \textbf{0.096$\pm$0.057} & 0.400$\pm$0.023 & 0.512$\pm$0.013 & 0.495$\pm$0.008 & 0.484$\pm$0.007 & 0.703$\pm$0.004 & 0.934$\pm$0.003 & 0.894$\pm$0.019 \\
    corel5k     & \textbf{0.002$\pm$0.001} & 0.331$\pm$0.009 & 0.289$\pm$0.132 & 0.280$\pm$0.009 & 0.601$\pm$0.008 & 0.331$\pm$0.011 & 0.400$\pm$0.009 & 0.531$\pm$0.014 \\
    rcv1subset1 & \textbf{0.005$\pm$0.001} & 0.109$\pm$0.005 & 0.138$\pm$0.013 & 0.118$\pm$0.004 & 0.228$\pm$0.009 & 0.318$\pm$0.008 & 0.459$\pm$0.010 & 0.707$\pm$0.011 \\
    Corel16k001 & \textbf{0.019$\pm$0.001} & 0.251$\pm$0.008 & 0.260$\pm$0.018 & 0.282$\pm$0.007 & 0.481$\pm$0.009 & 0.850$\pm$0.010 & 0.692$\pm$0.019 & 0.808$\pm$0.018 \\
    delicious   & \textbf{0.057$\pm$0.005} & 0.620$\pm$0.015 & 0.739$\pm$0.108 & 0.656$\pm$0.014 & 0.923$\pm$0.007 & 0.857$\pm$0.019 & 0.709$\pm$0.007 & 0.792$\pm$0.001 \\
    tmc2007     & \textbf{0.041$\pm$0.004} & 0.134$\pm$0.006 & 0.151$\pm$0.012 & 0.130$\pm$0.008 & 0.171$\pm$0.005 & 0.293$\pm$0.007 & 0.799$\pm$0.002 & 0.924$\pm$0.021 \\
    Image       & \textbf{0.143$\pm$0.008} & 0.190$\pm$0.056 & 0.303$\pm$0.093 & 0.283$\pm$0.044 & 0.278$\pm$0.010 & 0.185$\pm$0.013 & 0.802$\pm$0.018 & 0.899$\pm$0.007 \\
    iaprtc12    & \textbf{0.009$\pm$0.003} & 0.329$\pm$0.013 & 0.442$\pm$0.013 & 0.413$\pm$0.092 & 0.453$\pm$0.005 & 0.700$\pm$0.009 & 0.752$\pm$0.017 & 0.902$\pm$0.022 \\
    espgame     & \textbf{0.001$\pm$0.001} & 0.401$\pm$0.019 & 0.483$\pm$0.009 & 0.408$\pm$0.006 & 0.504$\pm$0.014 & 0.744$\pm$0.005 & 0.793$\pm$0.009 & 0.911$\pm$0.015 \\
    mirflickr   & \textbf{0.033$\pm$0.011} & 0.317$\pm$0.002 & 0.380$\pm$0.008 & 0.362$\pm$0.001 & 0.407$\pm$0.002 & 0.422$\pm$0.017 & 0.904$\pm$0.020 & 0.928$\pm$0.021 
    \end{tblr}
    \label{table:coverage_error}
}

\newcommand{\TableRankingLoss}{
    \caption{Ranking loss ($\downarrow$). Mean values and standard deviation over five runs are reported.}
    \begin{tblr}{
      hline{1,14} = {-}{0.08em},
      hline{2} = {-}{0.05em},
    }
    Datasets    & LAMC                   & SMILE                & AN          & WAN                  & ROLE        & GLOCAL      & MLML                 & DM2L        \\
    CAL500      & 0.373$\pm$0.005          & \textbf{0.238$\pm$0.009} & 0.276$\pm$0.123 & 0.250$\pm$0.008          & 0.390$\pm$0.009 & 0.303$\pm$0.012 & 0.422$\pm$0.010          & 0.492$\pm$0.020 \\
    scene       & 0.060$\pm$0.011          & 0.084$\pm$0.050          & 0.167$\pm$0.100 & \textbf{0.054$\pm$0.006} & 0.170$\pm$0.060 & 0.118$\pm$0.009 & 0.102$\pm$0.005	 & 0.402$\pm$0.040 \\
    yeast       & 0.219$\pm$0.017          & \textbf{0.160$\pm$0.003} & 0.170$\pm$0.010 & 0.165$\pm$0.002          & 0.172$\pm$0.010 & 0.340$\pm$0.010 & 0.342$\pm$0.020          & 0.512$\pm$0.012 \\
    corel5k     & \textbf{0.043$\pm$0.036} & 0.133$\pm$0.008          & 0.105$\pm$0.001 & 0.133$\pm$0.011          & 0.250$\pm$0.020 & 0.130$\pm$0.007 & 0.375$\pm$0.022          & 0.511$\pm$0.011 \\
    rcv1subset1 & \textbf{0.020$\pm$0.003} & 0.040$\pm$0.002          & 0.040$\pm$0.012 & 0.082$\pm$0.013          & 0.082$\pm$0.040 & 0.177$\pm$0.007 & 0.166$\pm$0.004          & 0.443$\pm$0.008 \\
    Corel16k001 & 0.161$\pm$0.010          & \textbf{0.129$\pm$0.003} & 0.140$\pm$0.018 & 0.144$\pm$0.008          & 0.239$\pm$0.015 & 0.708$\pm$0.011 & 0.301$\pm$0.003          & 0.488$\pm$0.018 \\
    delicious   & 0.139$\pm$0.002          & 0.127$\pm$0.010          & \textbf{0.092$\pm$0.003} & 0.098$\pm$0.008 & 0.340$\pm$0.028 & 0.505$\pm$0.018 & 0.334$\pm$0.001          & 0.532$\pm$0.009 \\
    tmc2007     & 0.045$\pm$0.007 		 & 0.048$\pm$0.001          & 0.049$\pm$0.001 & \textbf{0.039$\pm$0.005} & 0.098$\pm$0.013 & 0.131$\pm$0.006 & 0.144$\pm$0.004          & 0.509$\pm$0.080 \\
    Image       & \textbf{0.143$\pm$0.005} & 0.168$\pm$0.061          & 0.340$\pm$0.101 & 0.250$\pm$0.060          & 0.301$\pm$0.084 & 0.169$\pm$0.008 & 0.170$\pm$0.004          & 0.405$\pm$0.020 \\
    iaprtc12    & \textbf{0.080$\pm$0.017} & 0.119$\pm$0.001          & 0.182$\pm$0.113 & 0.155$\pm$0.055          & 0.177$\pm$0.022 & 0.409$\pm$0.008 & 0.300$\pm$0.021          & 0.453$\pm$0.009 \\
    espgame     & \textbf{0.052$\pm$0.037} & 0.149$\pm$0.011          & 0.155$\pm$0.016 & 0.118$\pm$0.011          & 0.256$\pm$0.008 & 0.470$\pm$0.003 & 0.322$\pm$0.031          & 0.409$\pm$0.019 \\
    mirflickr   & 0.155$\pm$0.021          & \textbf{0.115$\pm$0.002} & 0.120$\pm$0.005 & 0.134$\pm$0.012          & 0.170$\pm$0.010 & 0.201$\pm$0.013 & 0.875$\pm$0.008          & 0.457$\pm$0.014 
    \end{tblr}
    \label{table:ranking_loss}
}

\newcommand{\TableAveragePrecision}{
    \caption{Average precision ($\uparrow$). Mean values and standard deviation over five runs are reported.}
    \begin{tblr}{
      hline{1,14} = {-}{0.08em},
      hline{2} = {-}{0.05em},
    }
    Datasets    & LAMC                   & SMILE       & AN          & WAN         & ROLE        & GLOCAL      & MLML        & DM2L        \\
    CAL500      & \textbf{0.505$\pm$0.014} & 0.411$\pm$0.009 & 0.390$\pm$0.004 & 0.383$\pm$0.008 & 0.289$\pm$0.007 & 0.226$\pm$0.012 & 0.240$\pm$0.020 & 0.203$\pm$0.0010 \\
    scene       & \textbf{0.906$\pm$0.005} & 0.842$\pm$0.071 & 0.700$\pm$0.089 & 0.811$\pm$0.022 & 0.719$\pm$0.069 & 0.812$\pm$0.008 & 0.804$\pm$0.009 & 0.299$\pm$0.022 \\
    yeast       & \textbf{0.769$\pm$0.017} & 0.750$\pm$0.005 & 0.750$\pm$0.004 & 0.754$\pm$0.005 & 0.760$\pm$0.009 & 0.640$\pm$0.003 & 0.465$\pm$0.008 & 0.303$\pm$0.011 \\
    corel5k     & \textbf{0.742$\pm$0.013} & 0.310$\pm$0.006 & 0.301$\pm$0.010 & 0.289$\pm$0.011 & 0.206$\pm$0.014 & 0.228$\pm$0.004 & 0.101$\pm$0.011 & 0.088$\pm$0.013 \\
    rcv1subset1 & \textbf{0.748$\pm$0.031} & 0.622$\pm$0.002 & 0.606$\pm$0.008 & 0.609$\pm$0.004 & 0.581$\pm$0.004 & 0.233$\pm$0.005 & 0.231$\pm$0.008 & 0.093$\pm$0.008 \\
    Corel16k001 & \textbf{0.460$\pm$0.023} & 0.346$\pm$0.003 & 0.340$\pm$0.008 & 0.344$\pm$0.005 & 0.279$\pm$0.005 & 0.084$\pm$0.005 & 0.118$\pm$0.010 & 0.088$\pm$0.002 \\
    delicious   & \textbf{0.472$\pm$0.020} & 0.321$\pm$0.002 & 0.299$\pm$0.008 & 0.320$\pm$0.002 & 0.204$\pm$0.010 & 0.059$\pm$0.004 & 0.121$\pm$0.005 & 0.033$\pm$0.010 \\
    tmc2007     & \textbf{0.824$\pm$0.014} & 0.820$\pm$0.001 & 0.810$\pm$0.004 & 0.810$\pm$0.001 & 0.787$\pm$0.015 & 0.647$\pm$0.006 & 0.400$\pm$0.020 & 0.177$\pm$0.012 \\
    Image       & \textbf{0.809$\pm$0.004} & 0.786$\pm$0.048 & 0.610$\pm$0.080 & 0.695$\pm$0.048 & 0.696$\pm$0.045 & 0.777$\pm$0.008 & 0.665$\pm$0.017 & 0.300$\pm$0.013 \\
    iaprtc12    & \textbf{0.721$\pm$0.030} & 0.323$\pm$0.005 & 0.322$\pm$0.012 & 0.301$\pm$0.013 & 0.250$\pm$0.025 & 0.045$\pm$0.004 & 0.133$\pm$0.010 & 0.086$\pm$0.018 \\
    espgame     & \textbf{0.707$\pm$0.021} & 0.258$\pm$0.004 & 0.251$\pm$0.004 & 0.261$\pm$0.008 & 0.220$\pm$0.012 & 0.132$\pm$0.008 & 0.121$\pm$0.012 & 0.075$\pm$0.017 \\
    mirflickr   & \textbf{0.738$\pm$0.027} & 0.640$\pm$0.006 & 0.631$\pm$0.004 & 0.611$\pm$0.005 & 0.501$\pm$0.029 & 0.501$\pm$0.010 & 0.300$\pm$0.031 & 0.144$\pm$0.018 
    \end{tblr}
    \label{table:average_precision}
}

\newcommand{\TableDatasetStat}{
    \caption{Datasets detail}
    \begin{tblr}{
      column{2} = {r},
      column{3} = {r},
      column{4} = {r},
      column{5} = {r},
      cell{1}{6} = {r},
      hline{1,14} = {-}{0.08em},
      hline{2} = {-}{0.05em},
    }
    {Dataset\\Name} & {Number of\\Instances} & {Number of\\Features} & {Number of\\Labels} & {Cardinality} & {Domain} \\
    CAL500          & 502                    & 68                    & 174                 & 25.910                               & Music                 \\
    scene           & 2407                   & 294                   & 6                   & 1.067                                & Images                \\
    yeast           & 2417                   & 103                   & 14                  & 4.246                                & Biology               \\
    corel5k         & 5000                   & 499                   & 374                 & 3.550                                & Images                \\
    rcv1subset1     & 6000                   & 944                   & 101                 & 2.867                                & Text                  \\
    Corel16k001     & 13766                  & 500                   & 153                 & 2.863                                & Images                \\
    delicious       & 16105                  & 500                   & 983                 & 19.036                               & Text                  \\
    tmc2007         & 28596                  & 981                   & 22                  & 2.165                                & Text                  \\
    Image           & 2000                   & 294                   & 5                   & 1.208                                & Images                \\
    iaprtc12        & 19627                  & 1000                  & 291                 & 5.693                                & Images                \\
    espgame         & 20770                  & 1000                  & 268                 & 4.662                                & Images                \\
    mirflickr       & 25000                  & 1000                  & 38                  & 4.795                                & Images                
    \end{tblr}
    \label{table:dataset_stat}
}

%% file: IEEE-conference-template-062824.bbl
\begin{thebibliography}{10}
\providecommand{\url}[1]{#1}
\csname url@samestyle\endcsname
\providecommand{\newblock}{\relax}
\providecommand{\bibinfo}[2]{#2}
\providecommand{\BIBentrySTDinterwordspacing}{\spaceskip=0pt\relax}
\providecommand{\BIBentryALTinterwordstretchfactor}{4}
\providecommand{\BIBentryALTinterwordspacing}{\spaceskip=\fontdimen2\font plus
\BIBentryALTinterwordstretchfactor\fontdimen3\font minus \fontdimen4\font\relax}
\providecommand{\BIBforeignlanguage}[2]{{%
\expandafter\ifx\csname l@#1\endcsname\relax
\typeout{** WARNING: IEEEtran.bst: No hyphenation pattern has been}%
\typeout{** loaded for the language `#1'. Using the pattern for}%
\typeout{** the default language instead.}%
\else
\language=\csname l@#1\endcsname
\fi
#2}}
\providecommand{\BIBdecl}{\relax}
\BIBdecl

\bibitem{schapire2000boostexter}
R.~E. Schapire and Y.~Singer, ``Boostexter: A boosting-based system for text categorization,'' \emph{Machine Learning}, vol.~39, pp. 135--168, 2000.

\bibitem{zhang2013review}
M.-L. Zhang and Z.-H. Zhou, ``A review on multi-label learning algorithms,'' \emph{IEEE Transactions on Knowledge and Data Engineering}, vol.~26, no.~8, pp. 1819--1837, 2013.

\bibitem{wu2019tencent}
B.~Wu, W.~Chen, Y.~Fan, Y.~Zhang, J.~Hou, J.~Liu, and T.~Zhang, ``Tencent ml-images: A large-scale multi-label image database for visual representation learning,'' \emph{IEEE Access}, vol.~7, pp. 172\,683--172\,693, 2019.

\bibitem{deng2014scalable}
J.~Deng, O.~Russakovsky, J.~Krause, M.~S. Bernstein, A.~Berg, and L.~Fei-Fei, ``Scalable multi-label annotation,'' in \emph{Proceedings of the SIGCHI Conference on Human Factors in Computing Systems}, 2014, pp. 3099--3102.

\bibitem{cole2021multi}
E.~Cole, O.~Mac~Aodha, T.~Lorieul, P.~Perona, D.~Morris, and N.~Jojic, ``Multi-label learning from single positive labels,'' in \emph{Proceedings of the IEEE/CVF Conference on Computer Vision and Pattern Recognition}, 2021, pp. 933--942.

\bibitem{gibaja2015tutorial}
E.~Gibaja and S.~Ventura, ``A tutorial on multilabel learning,'' \emph{ACM Computing Surveys (CSUR)}, vol.~47, no.~3, pp. 1--38, 2015.

\bibitem{nam2014large}
J.~Nam, J.~Kim, E.~Loza~Menc{\'\i}a, I.~Gurevych, and J.~F{\"u}rnkranz, ``Large-scale multi-label text classification—revisiting neural networks,'' in \emph{Joint European Conference on Machine Learning and Knowledge Discovery in Databases}.\hskip 1em plus 0.5em minus 0.4em\relax Springer, 2014, pp. 437--452.

\bibitem{veit2017learning}
A.~Veit, N.~Alldrin, G.~Chechik, I.~Krasin, A.~Gupta, and S.~Belongie, ``Learning from noisy large-scale datasets with minimal supervision,'' in \emph{Proceedings of the IEEE Conference on Computer Vision and Pattern Recognition}, 2017, pp. 839--847.

\bibitem{xu2022one}
N.~Xu, C.~Qiao, J.~Lv, X.~Geng, and M.-L. Zhang, ``One positive label is sufficient: Single-positive multi-label learning with label enhancement,'' \emph{Advances in Neural Information Processing Systems}, vol.~35, pp. 21\,765--21\,776, 2022.

\bibitem{gao2025multi}
Y.~Gao, J.-Y. Zhu, M.~Xu, and M.-L. Zhang, ``Multi-label learning with multiple complementary labels,'' \emph{IEEE Transactions on Pattern Analysis and Machine Intelligence}, 2025.

\bibitem{zhu2017multi}
Y.~Zhu, J.~T. Kwok, and Z.-H. Zhou, ``Multi-label learning with global and local label correlation,'' \emph{IEEE Transactions on Knowledge and Data Engineering}, vol.~30, no.~6, pp. 1081--1094, 2017.

\bibitem{wu2014multi}
B.~Wu, Z.~Liu, S.~Wang, B.-G. Hu, and Q.~Ji, ``Multi-label learning with missing labels,'' in \emph{2014 22nd International conference on pattern recognition}.\hskip 1em plus 0.5em minus 0.4em\relax IEEE, 2014, pp. 1964--1968.

\bibitem{ma2021expand}
Z.~Ma and S.~Chen, ``Expand globally, shrink locally: Discriminant multi-label learning with missing labels,'' \emph{Pattern Recognition}, vol. 111, p. 107675, 2021.

\bibitem{pillai2011classification}
I.~Pillai, G.~Fumera, and F.~Roli, ``A classification approach with a reject option for multi-label problems,'' in \emph{Proceedings of the 16th International Conference on Image Analysis and Processing: Part I}, 2011, pp. 98--107.

\bibitem{pillai2013multi}
------, ``Multi-label classification with a reject option,'' \emph{Pattern Recognition}, vol.~46, no.~8, pp. 2256--2266, 2013.

\bibitem{wang2021can}
H.~Wang, W.~Liu, A.~Bocchieri, and Y.~Li, ``Can multi-label classification networks know what they don’t know?'' \emph{Advances in Neural Information Processing Systems}, vol.~34, pp. 29\,074--29\,087, 2021.

\bibitem{duarte2021plm}
K.~Duarte, Y.~Rawat, and M.~Shah, ``Plm: Partial label masking for imbalanced multi-label classification,'' in \emph{Proceedings of the IEEE/CVF Conference on Computer Vision and Pattern Recognition}, 2021, pp. 2739--2748.

\bibitem{chen2022extreme}
W.-T. Chen, Y.~Xia, and K.~Shinzato, ``Extreme multi-label classification with label masking for product attribute value extraction,'' in \emph{Proceedings of the Fifth Workshop on e-Commerce and NLP (ECNLP 5)}, 2022, pp. 134--140.

\bibitem{peng2023novel}
T.~Peng, Y.~Xue, J.~Li, and J.~Xu, ``A novel label selection algorithm based on principal component analysis and sparse approximation solution for multi-label classification,'' in \emph{2023 IEEE 35th International Conference on Tools with Artificial Intelligence (ICTAI)}.\hskip 1em plus 0.5em minus 0.4em\relax IEEE, 2023, pp. 532--537.

\bibitem{sadinle2019least}
M.~Sadinle, J.~Lei, and L.~Wasserman, ``Least ambiguous set-valued classifiers with bounded error levels,'' \emph{Journal of the American Statistical Association}, vol. 114, no. 525, pp. 223--234, 2019.

\bibitem{angelopoulos2023conformal}
A.~N. Angelopoulos, S.~Bates \emph{et~al.}, ``Conformal prediction: A gentle introduction,'' \emph{Foundations and Trends{\textregistered} in Machine Learning}, vol.~16, no.~4, pp. 494--591, 2023.

\end{thebibliography}
